\documentclass[conference]{IEEEtran}
\IEEEoverridecommandlockouts
\usepackage{cite}
\usepackage{amsmath,amssymb,amsfonts}
\usepackage{algorithm}
\usepackage[noend]{algpseudocode}
\usepackage{graphicx}
\usepackage{textcomp}
\usepackage{xcolor}
\usepackage{multirow}
\usepackage{multirow, array}
\usepackage{enumitem}
\usepackage{booktabs}
\def\BibTeX{{\rm B\kern-.05em{\sc i\kern-.025em b}\kern-.08em
    T\kern-.1667em\lower.7ex\hbox{E}\kern-.125emX}}
\begin{document}

\title{Multimodal Joint Emotion and Game Context Recognition in League of Legends Livestreams\\
\thanks{This work was supported by the EPSRC Centre for Doctoral Training in Intelligent Games \& Games Intelligence (IGGI) [EP/L015846/1] and the Digital Creativity Labs (digitalcreativity.ac.uk), jointly funded by EPSRC/AHRC/Innovate UK under grant no. EP/M023265/1.}
}

\author{\IEEEauthorblockN{Charles Ringer\IEEEauthorrefmark{1}\IEEEauthorrefmark{2},
James Alfred Walker\IEEEauthorrefmark{2},~\IEEEmembership{Senior Member, IEEE}, Mihalis A. Nicolaou\IEEEauthorrefmark{1}\IEEEauthorrefmark{3}}\\
\IEEEauthorblockA{\IEEEauthorrefmark{1}Department of Computing, Goldsmiths, University of London, London, UK SE14 6NW}
\IEEEauthorblockA{\IEEEauthorrefmark{2}Department of Computer Science, University of York, UK, York, UK YO10 5DD}
\IEEEauthorblockA{\IEEEauthorrefmark{3}Computation-based Science and Technology Research Center, The Cyprus Institute, Cyprus\\
Email: c.ringer@gold.ac.uk,
james.walker@york.ac.uk,
m.nicolaou@cyi.ac.cy}
}

 \IEEEpubid{\begin{minipage}{\textwidth}\ \\[12pt]
978-1-7281-1884-0/19/\$31.00 \copyright 2019 IEEE
\end{minipage}}

\maketitle

\begin{abstract}
Video game streaming provides the viewer with a rich set of audio-visual data, conveying information both with regards to the game itself, through game footage and audio, as well as the streamer's emotional state and behaviour via webcam footage and audio.  Analysing player behaviour and discovering correlations with game context is crucial for modelling and understanding important aspects of livestreams, but comes with a significant set of challenges - such as fusing multimodal data captured by different sensors in uncontrolled  (`in-the-wild') conditions.  Firstly, we present, to our knowledge, the first data set of \textit{League of Legends} livestreams, annotated for both streamer affect and game context. Secondly, we propose a method that exploits tensor decompositions for high-order fusion of multimodal representations.  The proposed method is evaluated on the problem of jointly predicting game context and player affect, compared with a set of baseline fusion approaches such as late and early fusion. Data and code are available at https://github.com/charlieringer/LoLEmoGameRecognition
\end{abstract}


\section{Introduction} \label{sec:intro}
\begin{figure}[t]
\centering
\includegraphics[width=0.5\textwidth]{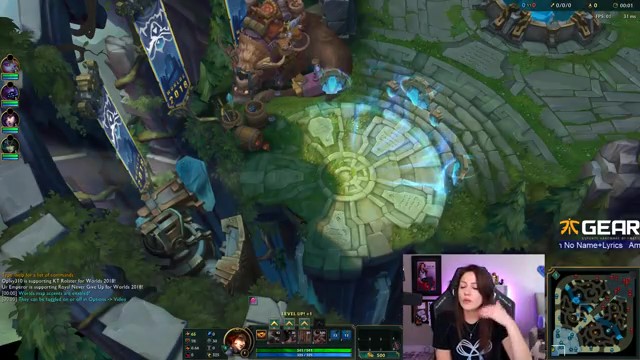}
\caption{Example screenshot from a \textit{League of Legends} livestream.}
\label{fig:lol}
\end{figure}

Livestreaming is an exciting and emerging area of video games entertainment. People find watching other people play games compelling \cite{Kaytoue2012}, as evidenced by the popularity of streaming services like \texttt{TWITCH.TV}\footnote{www.twitch.tv}. Typically, a live stream entails the broadcast of a set of both visual and auditory data.  This includes game footage along with a webcam overlay showing the streamer (Fig. \ref{fig:lol}), as well as auditory data which includes in-game audio as well as speech and non-verbal cues. As a result, livestreaming presents a rare opportunity for studying streamer emotion and affect at the same time as the stimulus for this emotion, i.e. their game-play experience. 

Modelling livestreams is an inherently `in-the-wild' paradigm, using organically generated real-world data \cite{10.1007/978-3-319-99978-4_2, streambeh18}, and therefore subject to many complicating factors such as visual and audio occlusions. Such occlusions can either be temporary, e.g. the streamer looking away from the camera, or permanent, e.g. overlays placed over the game scene or the music that the streamer is listening. Other difficulties include streamers having their webcams at a range of angles, using varying levels of lighting, and choosing different volume levels between their voice, music and the game audio. Additionally, modelling streamer affect and game context is a multimodal, or multi-view, problem where a key challenge is joining information from multiple views, e.g. webcam, game footage and audio, to form a single model. It is possible that these complicating factors contribute to the lack of past study into audio-visual stream data, but we feel that despite this there are compelling reasons to study livestreaming. In fact, one goal of this work is to invigorate the research community and spark interest in this topic.

This paper presents two contributions. Firstly, a data set of streamers playing the popular Multiplayer Online Battle Arena (MOBA) game \textit{League of Legends}\footnote{Riot Games, 2009} is presented, along with annotations for both streamer affect and in-game context. Secondly, a novel method for fusing multiple views by modelling high-order interactions using a `Tensor Train layer'\cite{Yang2017} is presented and evaluated on this dataset. Additionally, we present a comparison of this method with several existing fusion approaches, thus providing baseline results for the dataset.

In this work, experiments are presented for all fusion methods in terms of modelling affect and game context both jointly as well as separately.  In essence, this paper acts as a platform inviting further research into this problem, and a starting point for the study of supervised learning in terms of modelling the multiple facets of livestreams and their interactions.

%In Section 2 a discussion of previous works is carried out, Section 3 details a data set of League of Legends streamers. Section 4 discussion the architecture for the models experimented with, and the details of this experimental procedure are presented in Section 5. Finally, Section 6 discusses our results and section 7 contains conclusions.

\section{Related Work}
\subsection{Analysis of Game Streams}
Livestreaming is a young technology and so currently lacks a wealth of former work dedicated to it. Existing studies into modelling livestreams have focused on detecting `highlights', i.e exciting or important moments, in streams. Chu \textit{et al.} \cite{Chu2015, Chu2017} looked into various facets of \textit{League of Legends} esports broadcasts, building models of highlight detection, focusing on modelling hand-crafted features such as the number of players on screen, and event detection, utilising text recognition on in-game messages. Additionally, our previous work \cite{Ringer} focused on using unsupervised learning to detect highlights in livestreams of \textit{Player Unknown's Battlegrounds} using a technique similar to `feature fusion', as presented in this paper, to fuse measures of novelty across views over time. 

Often, instead of studying streams themselves, past research has focused on the social and community aspects of livestreaming. Examples include Smith \textit{et al.} \cite{Smith2013}, Recktenwald \cite{RECKTENWALD201768} and, Robinson \textit{et al.} \cite{Robinson2017} who studied streamer - viewer interactions. Others have examined the behaviour of streams in general, taking a higher level view by looking at features such as the number of viewers and the length of streams, such as Kaytoue \textit{et al.} \cite{Kaytoue2012} and Nascimento \textit{et. al} \cite{7000165}.

\subsection{Studies of Player Experience Through Visual and Audio-Visual Data}
Most prior work into audio-visual models of player experience do not use livestreaming platforms as their source of data. For example, the Platform Experience Dataset \cite{7344647} is a data set of players playing the platformer game \textit{Infinite Mario Bros}. This data set has been utilized in several studies, for example, Shaker \textit{et al.} \cite{Shaker2013} developed player experience modelling techniques, while Asteriadis \textit{et al.} \cite{Asteriadis2012} used this data set to cluster player types. Additionally, off-the-shelf affect models have been used to study player experience, for example Blom \textit{et al.} \cite{Blom} explored how these models could be used to personalise game content and Tan \textit{et al.} \cite{Tan2012} performed a feasibility study exploring player experience modelling via visual cues, concluding that facial analysis is a rich data source which warrants more exploration.

\subsection{Multimodal Machine Learning}
The problem of analysing multiple views is an open challenge in the affective computing and computer vision communities and describe situations where multiple data sources, e.g. audio and images, depict different views of the same event. Generally speaking, attempts to tackle this problem have focused on the question of fusion - how can multiple views be joined, or fused, into a single model. There is no general consensus on what the most appropriate fusion technique is with many studies attempting to perform fusion in different ways within the network architecture \cite{Katsaggelos2015, Poria2017}. Some of the most popular techniques for fusion in `end-to-end' systems, where the inputs are raw data extracted from the video and the outputs are the classifications, include early (feature) fusion and late (decision-level) fusion. While early fusion refers to concatenating raw features or representations (e.g. \cite{Ngiam, WOLLMER2013153}), late fusion is usually performed on bottleneck features (e.g. \cite{6865245}). Note that other fusion approaches exist, e.g. model fusion, where separate models are utilized for each view and subsequently aggregated (e.g. EmoNets \cite{Kahou2015}). However, such approaches are not naturally suited to an end-to-end system as two stages are often required during training.

A significant advantage of utilizing multimodal data, as Ngaim et al. \cite{Ngiam} note, is that exploiting information from multiple views can significantly aid learning from noisy and imperfect data (e.g. when audio is noisy).  This suggests that multimodal fusion may be well suited to the problem at hand, where noise is introduced due to uncontrolled conditions (discussed in Section \ref{sec:intro}).  This is also crucial for modelling behaviour and affect, as it is well known that audio information is more suited to predicting emotional arousal, while visual data is more suited for modelling valence \cite{Tzirakis}.  Therefore, fusing these views allows for holistic modelling of affect.  We refer the interested reader to \cite{Katsaggelos2015} and \cite{Poria2017} for more details on the literature in multimodal learning.

\subsection{Tensor Decompositions}
This paper presents a method for utilizing tensor decompositions as a fusion mechanism in order to model high order relationships between multiple views. For the purposes of this work, a tensor refers to the general term for an array of values where the rank refers to the dimensionality of the array e.g. a vector is a rank-1 tensor and a matrix is a rank-2 tensor. The proposed approach is similar to Zadeh \textit{et al.} \cite{Zadeh2017}, where Tensor Fusion was employed to fuse multiple views in a sentiment analysis task. Likewise, this work utilises a Tensor Train\cite{doi:10.1137/090752286} \cite{Novikov} layer that decomposes a tensor into a set of simpler and smaller ones in a weight-efficient manner.
%a method for decomposing a tensor into a set of smaller tensors, which replaces a dense layer and is able to take this multimodal tensor and model it in a weight-efficient manner. 
Other researchers have used similar approaches to aid different tasks, e.g. Yang \textit{et al.} \cite{Yang2017} utilizes a (recurrent) Tensor Train layer, directly on the pixels of a video replacing the convolutional layers. Kossaifi \textit{et al.} \cite{Kossaifi} used similar tensor methods on the unflattened output of a Convolutional Neural Network and showed that these types of decomposition can be trained in an end-to-end model. Anandkumar \textit{et al.} \cite{Anandkumar2014} provide a comprehensive overview of tensor decompositions for learning latent variable models. 

\section{Data Set}
\textit{League of Legends} is a MOBA game where two teams of five players compete with the goal of reaching and destroying the opposing team's base. It is one of the most popular esports, with at least 5 professional leagues and numerous global competitions\footnote{https://eu.lolesports.com/}. In addition, it is an incredibly popular game for streamers and is regularly in the top 3 most popular games being streamed on \texttt{TWITCH.TV}\footnote{https://twitchstats.net/}. Video data\footnote{Code for all models along with the data set can be downloaded from https://github.com/charlieringer/LoLEmoGameRecognition} was gathered from streamers playing \textit{League of Legends} on \texttt{TWITCH.TV}. The data set consists of 10 streamers, five male and five female, streaming in English, and using \texttt{TWITCH.TV}. 20 minutes of footage was gathered from 3 games for each streamer for a total of 10 hours of footage. This data was then segmented into five second long non-overlapping segments, for a total of 7200 video clips. Each clip was manually annotated across three labels, two of which related to the streamers affect and one of which related to what the streamer was doing in the game. Note that while the number of streamers is limited by the effort required to manually annotated data the length of the data set is in keeping with other works e.g. \cite{Chu2015, Chu2017}.

\begin{table*}[]
\begin{center}
\caption{Distribution of annotations in the raw data set.}\label{table:annodistro}
\begin{tabular}{@{}rrrrrrrrrrrrrr@{}}
\toprule
\multicolumn{3}{c}{Valence} & \multicolumn{2}{c}{Arousal} & \multicolumn{9}{c}{Game Context}\\
\cmidrule(r){1-3}\cmidrule(lr){4-5}\cmidrule(l){6-14}
Neg & Neut & Pos & Neut & Pos & In Lane & Shopping & Ret. to Lane & Roaming & Fighting & Pushing & Defending & Dead & Misc. \\
\midrule
246 & 6,227 & 727 & 6,755 & 445 & 2,418 & 294 & 591 & 1,422 & 892 & 213 & 233 & 831 & 308 \\
\bottomrule
\end{tabular}
\end{center}
\end{table*}

\subsection{Emotional Affect Annotation}
Each clip was annotated for affect, using the streamer's facial, bodily and vocal cues to judge their emotional state. Affect was annotated across two dimensions, valence and arousal. `Valence' relates to the positive/negative axis of emotion whereas `arousal' relates to how strongly someone is feeling/displaying emotion. For valence each clip was rated on a three-point scale, positive, neutral or negative. For arousal, a two-point scale, neutral or positive, was used because video games, especially \textit{League of Legends}, are not generally designed to elicit negative arousal and so this did not appear often in the data set. Therefore each clip receives a valance and arousal classification according to observable displays of the following:

\begin{description}
\item [Negative Valence] Negative feeling e.g sadness.
\item [Neutral Valence] A lack of discernible valence.
\item [Positive Valence] Positive feeling e.g. happiness.
\item [Neutral Arousal] A lack of discernible arousal. 
\item [Positive Arousal] Strong emotional response e.g. anger or excitement.
\end{description}

As can be seen from Table \ref{table:annodistro} there is a huge imbalance between classes with a skew towards neutral affect in both dimensions. This is to be expected; often gamers are engrossed in gameplay, and as a result, do not show outward emotion. This adds another complicating factor to the difficulty of learning affect in a livestream setting.

\subsection{Game Annotation}
\looseness-1Each clip was also annotated for game context, relating to what the streamer was doing during the clip. This behaviour is not always represented on screen for the full duration of the clip, in most cases due to players quickly switching their camera in-game to observe what others are doing.  %\textcolor{red}{Still, the annotation represents the majority of frames in the video.}
%. This is due to players often quickly switch their camera view to check what other players are doing. 
Eight categories were chosen which represent the majority of gameplay and for the occasions where the player is doing something outside of the scope of the categories a ninth `miscellaneous' category was used: 

\begin{description}
\item [In Lane] Farming `creeps' (game controlled enemies) in a lane. Often the default action.
\item [Shopping] Spending gold earned in-game on items.
\item [Returning to Lane] Walking back to lane after re-spawning, shopping or returning to base for health. 
\item [Roaming] Roaming the `jungle' area, the space between lanes.
\item [Fighting] Engaging in player vs player combat. 
\item [Pushing] Pushing into and attacking the enemy base.
\item [Defending] Defending their own base.
\item [Dead] Killed and is awaiting re-spawn.
\item [Miscellaneous] Something not covered above.
\end{description}

\looseness-1Similarly to the affect annotations, there is an imbalance between game event classes, although less pronounced. `In Lane' and `Roaming' are the most popular activities, representing 33.58\% and 19.75\% of the data respectively, shown in Table \ref{table:annodistro}.

\subsection{Data Pre-Processing and Over-Sampling}
\looseness-1Once annotated, all clips where the game annotation was `miscellaneous' were removed because this label does not accurately represent the content of the clip. Next, the data set was split randomly into 20\% testing data, 1375 clips, and 80\% training data, 5517 clips. The training data underwent a further pre-possessing step, oversampling, to help address the class imbalance in the data. Traditionally, oversampling algorithms, e.g. SMOTE \cite{Chawla} and ADASYN \cite{4633969}, generate synthetic data for minority classes by finding points between two existing minority examples. However, these techniques are not applicable to videos, which features both incredibly high dimensionality and important inter-view relationships. Therefore, synthetic data is generated as clones of existing minority class data. 

\looseness-1It is important to ensure that oversampling a minority class in one output does not increase the majority class if it belongs to a different output. Therefore, the least and most represented classes across all annotations are calculated, with a weighting applied to account for the varying number of classes between outputs, see Equation \ref{eq:repweighting} where $w_c$ is the representation weight for a class $c$, $T_c$ is the total for this class, $T_d$ is the total data points in the data set, and $N_c$ is the number of classes for this output. Next, a data point is selected at random which is in the least represented class and not in the most represented class. The selected clip is then cloned in the data set. This process is repeated until either a predefined threshold is reached (the chosen threshold for this work was the size of the initial data-set) or no data satisfies the selection requirement. The result of this oversampling is shown in Table \ref{table:impact}.

\begin{equation} \label{eq:repweighting}
w_{c} = \frac{T_{c}}{T_{d}/(1/N_{c})}
\end{equation}

\begin{table*}[]
\begin{center}
\caption{Impact of the Oversampling Technique on the Training Data Set.}\label{table:impact}
\begin{tabular}{@{}lrrrrrrrrrrrrr@{}}
\toprule
& \multicolumn{3}{c}{Valence} & \multicolumn{2}{c}{Arousal} & \multicolumn{8}{c}{Game Context}\\
\cmidrule(lr){2-4}\cmidrule(lr){5-6}\cmidrule(l){7-14}
& Neg & Neut & Pos & Neut & Pos & In Lane & Shopping & Ret. to Lane & Roaming & Fighting & Pushing & Defending & Dead  \\
\midrule
Before & 0.033 & 0.862  & 0.104  & 0.937 & 0.063 & 0.35 & 0.043 & 0.084 &  0.206  & 0.13 & 0.03  & 0.033   & 0.123 \\
\midrule
After & 0.259 & 0.483  & 0.257  & 0.725 & 0.275 & 0.181 & 0.097 & 0.097 & 0.146 & 0.103 & 0.181  & 0.097  & 0.181 \\
\bottomrule
\end{tabular}
\end{center}
\end{table*}

\looseness-1The data is also processed by taking each five-second clip and extracting visual and audio frames at a rate of four frames per second. Game images are $128 \times 128 \times 3$ down-sampled images taken from the frame. They represent the game context on the screen and have a black patch placed over the streamer's webcam. The streamer's webcam images are $64 \times 64 \times3$ down-sampled images taken from the frame and represent what is present in the streamer's webcam and contain no gameplay data. The audio represents a joint stream, containing the streamer's voice, the game audio, along with occlusions such as any music the streamer is listening to and represents the raw audio waveform as a single vector of length $5512$. Therefore, the input data for each clip consists of 20 frames ($4$ fps $\times$ $5$ seconds) represented as two image tensors and one audio vector. The temporal and spacial down-sampling was chosen empirically to provide a reasonable middle ground between representing the original clip and reducing the data passed into the network for performance reasons.

\begin{figure*}[t]
\centering
\includegraphics[width=\textwidth]{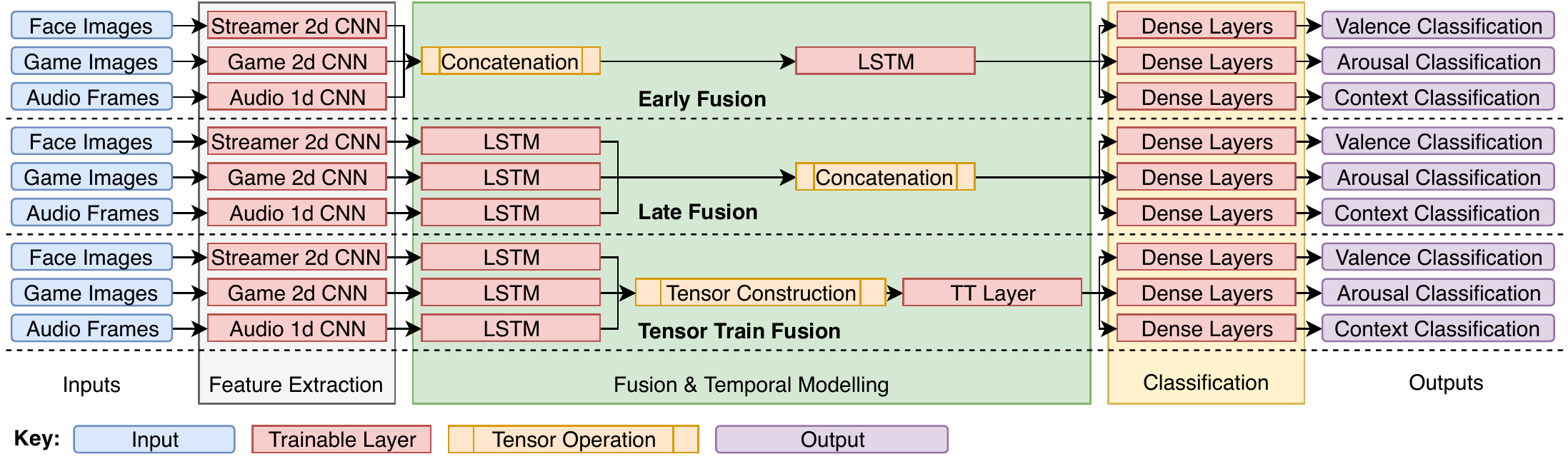}
\caption{Comparison between the high level architectures of the three fusion techniques presented in this paper. All modes use the same feature extraction and classification layer shapes but differ in how the input views are fused. Best viewed on a computer.}
\label{fig:models}
\end{figure*}

\section{Methodology}\label{sec:meth}
\looseness-1Three models with a similar underlying structure were developed to compare fusion technique. First, a set of latent features are extracted from each frame using a set of convolutional neural networks (CNNs). When considering views containing image data a 2D CNN is used, whereas on the audio stream a 1D CNN is applied, due to the one-dimensional shape of the audio data. These features are then modelled temporally using several Long Short-Term Memory (LSTM) \cite{lstm} recurrent layers. Finally, several fully connected layers are used to extract a set of classifications, dependant on the task presented to the network. The difference between models, discussed in \ref{ssec:mvf}, lies in how the multimodal latent representations after the feature extraction stage are fused to build a shared representation of the input. The high-level architecture for each model is demonstrated in Fig. \ref{fig:models}.

\subsection{Feature Extraction}
\begin{figure}[t]
\centering
\includegraphics[width=0.5\textwidth]{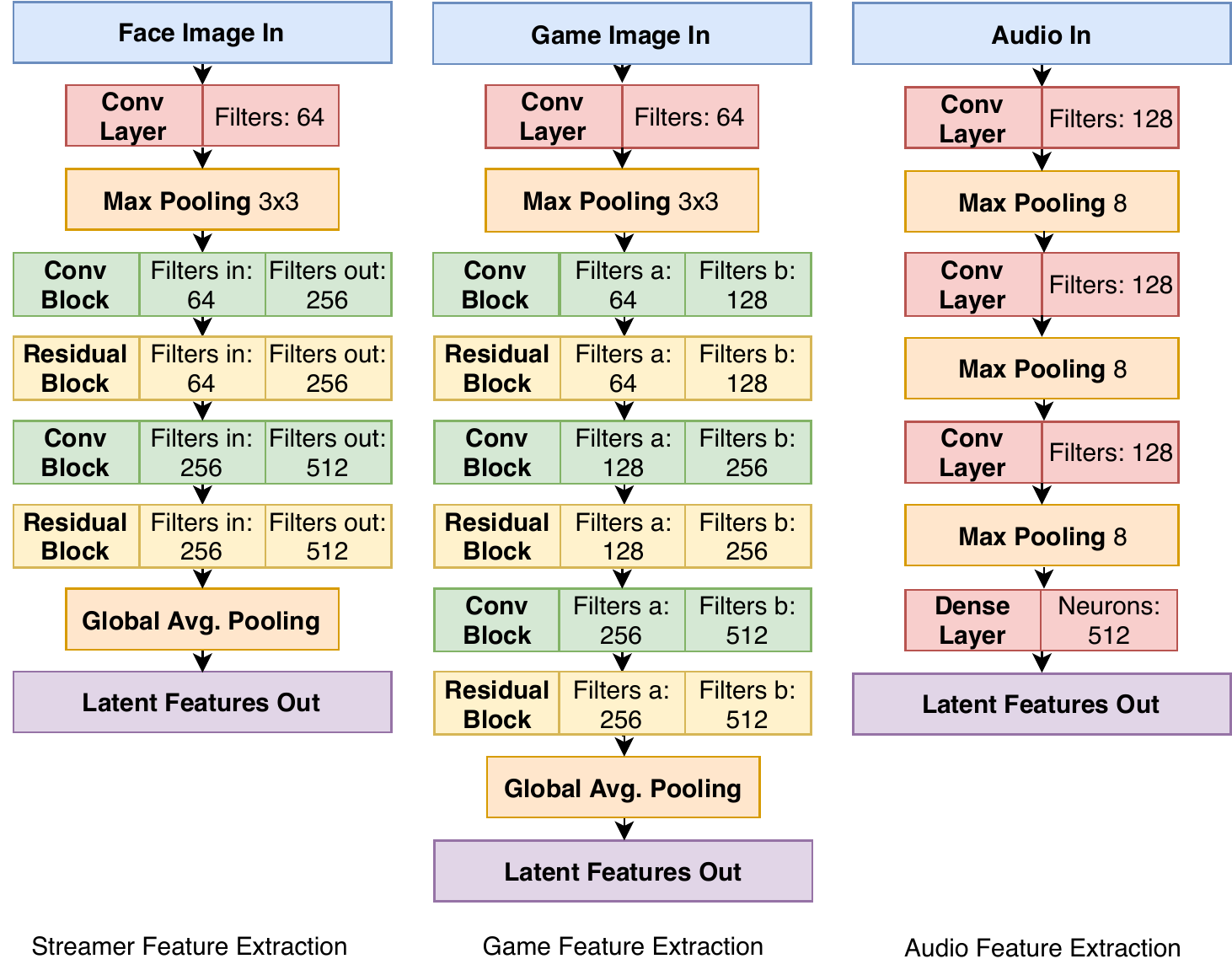}
\caption{Architectures of feature extraction modules. Note that each `Conv Block' and `Residual Block' are in turn multiple layers, resulting in a deeper network than shown. The makeup of these blocks is show in Fig. \ref{fig:res_conv}.}
\label{fig:feat_extract}
\end{figure}

\begin{figure}[t]
\centering
\includegraphics[width=0.5\textwidth]{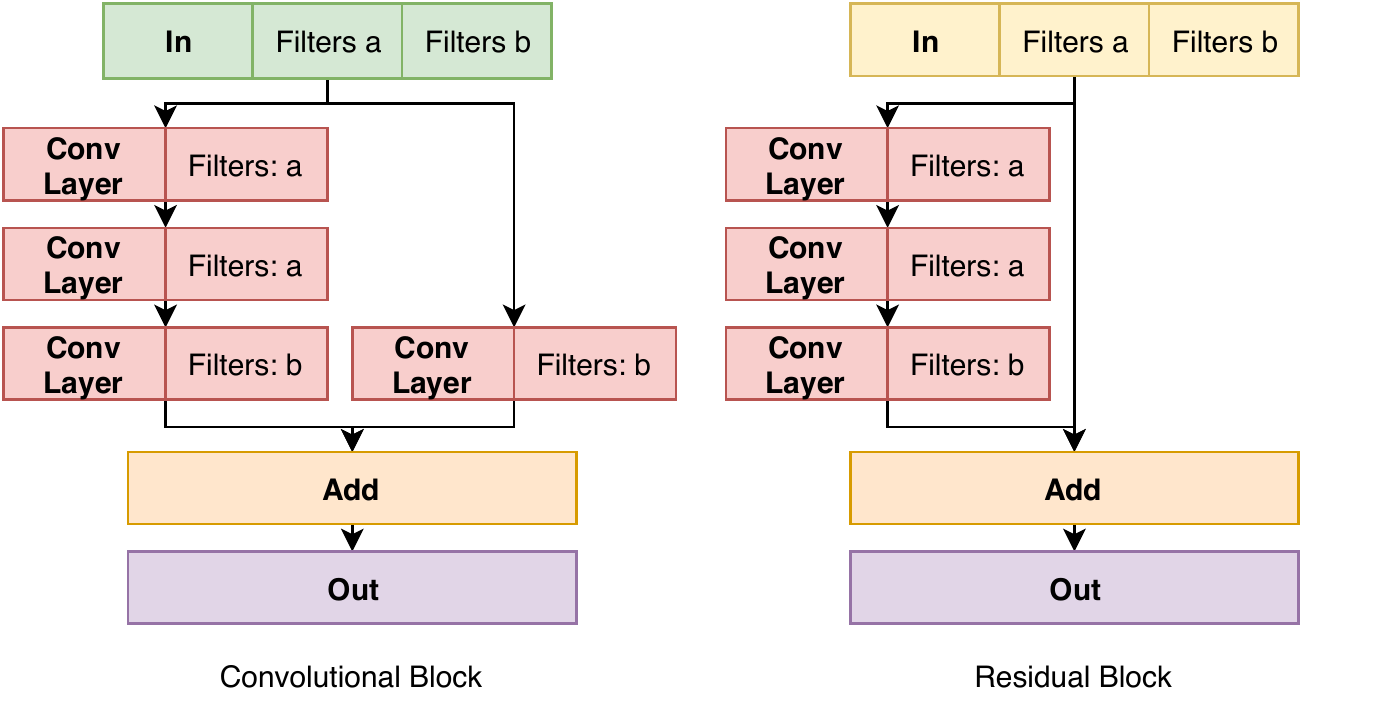}
\caption{Structure of the convolutional and residual blocks.}
\label{fig:res_conv}
\end{figure}

\looseness-1All models use the same set of CNN architectures for extracting a vector of latent features from each view. The two image networks, for streamer and game data, use a 2D CNN with a series of residual blocks \cite{7780459} to aid in back-propagating gradient. The difference being that the streamer network expects a smaller input image and as a result requires fewer layers and weights to satisfactorily model the features required. For modelling audio, a feature extractor with 3 1D convolutional layers followed by a dense layer was used. After each convolutional layer across all feature extractors, ReLU activation and Batch Normalization are applied. A visual layout of these architectures can be found in Fig. \ref{fig:feat_extract}. Convolutional and residual block structures are illustrated in Fig. \ref{fig:res_conv}. For more details on residual CNNs, see \cite{7780459}.

\begin{figure}[t]
\centering
\includegraphics[width=0.5\textwidth]{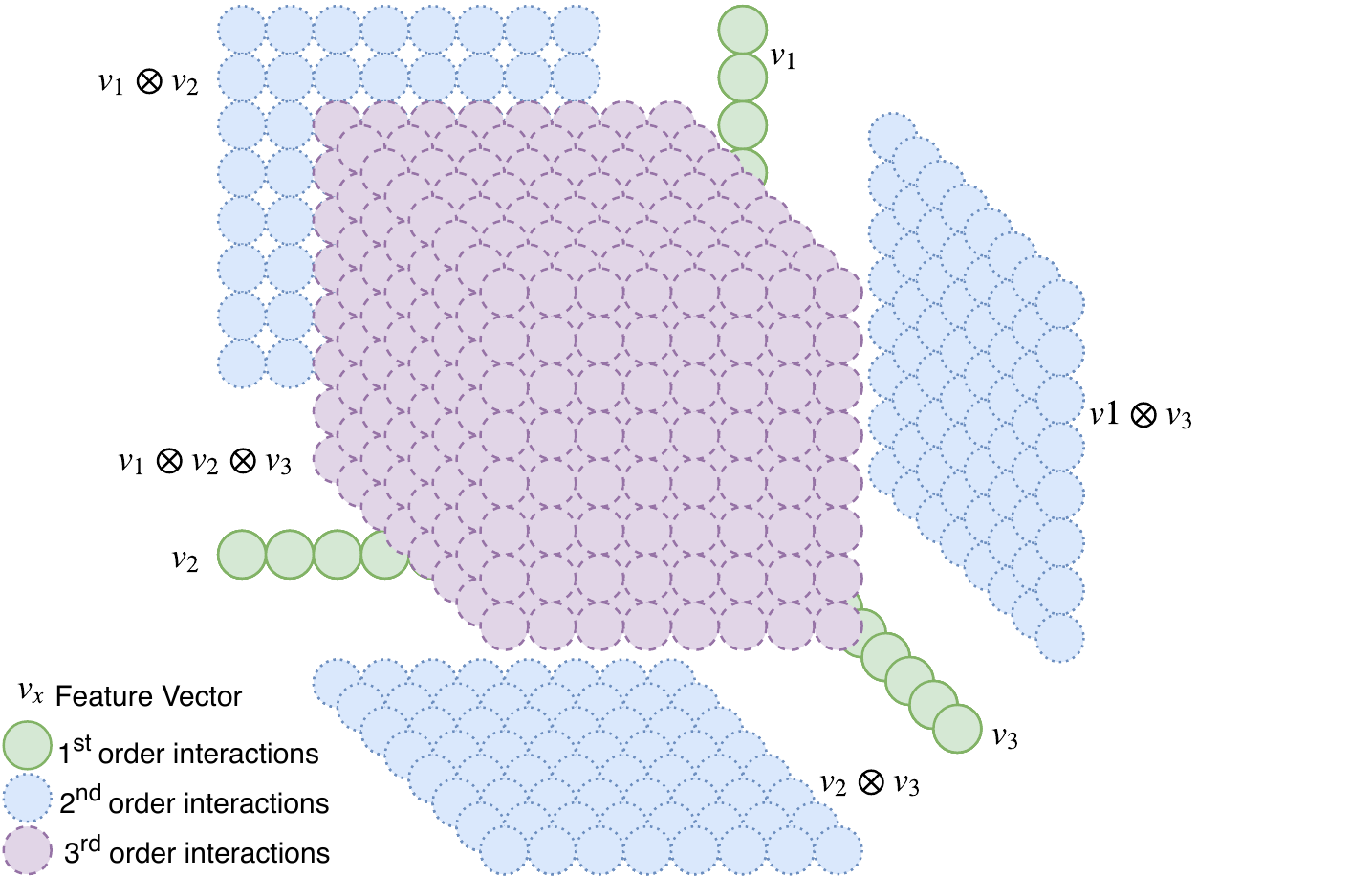}
\caption{Exploded structure of the 1\textsuperscript{st}, 2\textsuperscript{nd} and 3\textsuperscript{rd} order interaction tensor constructed before the Tensor Train layer. Note: Each feature here represents 16 features in the model. Best viewed on a computer.}
\label{fig:tensorconst}
\end{figure}

\subsection{Multimodal Fusion and Temporal Modelling}\label{ssec:mvf}
\looseness-1Once latent features have been extracted frame by frame from the various views, the next step is to fuse the views together to form a single view in addition to modelling per-frame features temporally. The purpose of this process is to extract a single feature vector corresponding to a joint representation across both views and time. Traditionally, concatenating input or representation space vectors has been one of the most popular approaches \cite{Katsaggelos2015}.  Nevertheless, this method can fail in terms of modelling interactions between views. The fusion approaches detailed in this work include both early and late fusion, as well as a novel method based on tensor decompositions to facilitate modelling high-order interactions between views in an efficient manner. The methods are detailed in the following sections.

\subsubsection{Early Fusion}
\looseness-1In the early fusion model, fusion is performed by taking the 512 features extracted for each view per frame and fusing them into a single vector of $l = 1536$. A stack of 20 feature vectors, one for each frame, is then fed into 2 LSTM layers each with 384 neurons. Before and after the LSTM layers, batch normalization and 20\% drop out are applied to guard against over-fitting and aid generalisation. The resulting vector represents the fused representation of this clip across views and time. 

\subsubsection{Late Fusion}
\looseness-1Late fusion is implemented with separate LSTM layers, with 128 neurons per layer for each view. Batch Normalization and Dropout layers are applied before and after the layers. As such, a single feature vector for each view represents the latent representation of the entire clip across one view. Concatenation of the various views then occurs after the LSTM step, and right before classification.

\subsubsection{TensorTrain Fusion}
\looseness-1Both early and late fusion models concatenate feature vectors from each view into a single vector. However, this fusion approach has no explicit representations which capture the relationship between variables in different views. To better capture interactions between different views, we construct a tensor that models up to third order interactions between views.  That is, given feature vectors  $v_x$,  $v_y$,  $v_z$ each corresponding to separate views, % are expanded to $[1, v_x]$, $[1, v_y]$, $[1, v_z]$. 
we take the cross product as

\begin{equation}
    z = \begin{bmatrix}1 \\ v_x\end{bmatrix} \otimes \begin{bmatrix}1 \\ v_y\end{bmatrix} \otimes \begin{bmatrix}1 \\v_z\end{bmatrix} \\ 
\end{equation}
%\begin{center}
    $were$
%\end{center}
\begin{equation*}
    \begin{bmatrix}1 \\ v_x\end{bmatrix} \otimes \begin{bmatrix}1 \\ v_y\end{bmatrix} = \begin{bmatrix}1 & v_x \\ v_y & v_x \otimes v_y\end{bmatrix}
\end{equation*}
As such the fused feature tensor $z \in \mathbb{R}^{129, 129, 129}$ is created which contains $v_x$, $v_y$, $v_z$, $v_x \otimes v_y$, $v_x \otimes v_z$, $v_y \otimes v_z$, and $v_x \otimes v_y \otimes v_z$. The resulting tensor is shown in Fig. \ref{fig:tensorconst}. 

\looseness-1At this point, it would be possible to flatten this feature tensor and then pass it through a series of dense classification layers. However, due to the size of the tensor ($2,146,689$ elements), this is computationally infeasible. Connecting this tensor to a fully connected layer with 128 neurons would result in $128 \times 2,146,689 = 274,776,192$ weights for this layer alone. Therefore, a Tensor Train layer \cite{Yang2017} is used to connect the latent tensor with the classification layers. This layer replaces a dense layer and represents its weight matrix as a series of smaller tensors, the tensor cores. In this case, a tensor train with ranks (1, 2, 4, 4, 2, 1) is used. Each element in the weight tensor is then approximately represented as a product of these tensors thus allowing the model to learn the weighted mapping between the input tensor $z$ and the output vector with far fewer parameters, around 11,000, resulting in a space saving of $4 \times 10^{-5}$. The Tensor Train layer extracts 384 features, the same number features of concatenating the original feature vectors, which are then passed forward to the classification layers. This approach thus incurs only a small increase in weights whilst in theory modelling these important higher order interactions.

\subsection{Classification}
\looseness-1The fused feature vector is then passed into several dense layers to perform classification. These layers act as separate task specific `heads'. For each task, first, the feature vector is passed through a 128 neuron dense layer with ReLU activation before the final classification layer, which is calculated by taking the softmax across $n$ neurons where $n$ = the number of classes (e.g. for Valence $n$ = 3). 

\section{Experiment}
\looseness-1Each model presented in Section \ref{sec:meth} was trained on three tasks. Firstly, to learn a joint representation of both game context and streamer affect, and thus classify valence, arousal and game context simultaneously. Secondly, the task of only learning the game context. Finally, the task was learning just the streamer's affect. 

\looseness-1We used Keras \cite{chollet2015keras} with the Tensorflow backend for implementing each model. Models were optimized with ADAM \cite{article} where $\alpha = 0.0005$. For each learning task, each network was trained for 100 epochs using an NVIDIA GTX 1080 GPU. The number of weights per model can be found in Table \ref{table:params}. For each output, a set of class weights were implemented as an additional measure to tackle the bias in the data set. To calculate class weights for each class $x$, an initial weight $i_x$ is calculated then a scaled weight $w_x$ is calculated so that all weights for an output sum to one:

\begin{equation}
    i_x = \frac{T_{d}}{T_{x} \times N_{c}}
\end{equation}
\begin{equation}
    w_x, w_y, ... = \frac{i_x}{sum(i_x, i_y, ...)}, \frac{i_y}{sum(i_x, i_y, ...)},  ... 
\end{equation}
Where $T_{d}$ is the total data points in the data set, $T_x$ is the total data points for class $x$ and $N_c$ is the number of classes for this output, e.g. for a valence class $N_c = 3$ as there are 3 possible valence classifications.

\begin{table}[]
\begin{center}
\caption{Comparison of Trainable Weights Across Models.}\label{table:params}
\begin{tabular}{@{}lrrr@{}}
\toprule
 Task   & Early Fusion & Late Fusion & Tensor Train Fusion \\ \midrule
Affect + Game & 9,382,029      & 6,629,517   & 6,640,939           \\
Affect        & 9,331,717      & 6,579,205   & 6,590,627           \\
Game          & 9,282,824      & 6,530,312   & 6,541,734         \\ \bottomrule
\end{tabular}
\end{center}
\end{table}

\section{Results}
\begin{table*}[]
\caption{F1 Scores for Each Label Across All Models. For each model F1 scores for each task and each label are reported. Best result for each class in \textbf{bold}.}\label{table:prf1}
\resizebox{\textwidth}{!}{%
\begin{tabular}{@{}llrrrrrrrrrrrrr@{}}
\toprule
Model                                                                      & Task   & Neg V & Neut V & Pos V & Neut A & Pos A & In Lane & Shopping & Returning & Roaming & Fighting & Pushing & Defending & Dead  \\ \midrule
\multirow{2}{*}{\begin{tabular}[c]{@{}l@{}}Early \\ Fusion\end{tabular}} & Joint  & 0.194 & 0.911  & \textbf{0.362} & 0.969  & 0.509 & 0.778   & 0.797    & 0.496     & 0.667   & 0.515    & 0.568   & 0.544     & 0.899 \\
                                                                           & Single & 0.206 & 0.905  & 0.345 & 0.966  & \textbf{0.540} & 0.842   & 0.724    & 0.591     & 0.794   & 0.565    & 0.610   & \textbf{0.667}     & 0.924 \\ \midrule
\multirow{2}{*}{\begin{tabular}[c]{@{}l@{}}Late\\ Fusion\end{tabular}}     & Joint  & \textbf{0.286} & 0.925  & 0.297 & 0.964  & 0.465 & 0.840   & \textbf{0.828}    & \textbf{0.615}     & 0.805   & \textbf{0.635}    & \textbf{0.652}   & 0.557     & 0.906 \\
                                                                           & Single & 0.088 & 0.918  & 0.340 & 0.968  & 0.491 & 0.791   & 0.776    & 0.513     & 0.774   & 0.581    & 0.582   & 0.452     & 0.937 \\ \midrule
    \multirow{2}{*}{\begin{tabular}[c]{@{}l@{}}TT \\ Fusion\end{tabular}}      & Joint  & 0.102 & 0.926  & 0.325 & \textbf{0.971}  & 0.476 & \textbf{0.848}   & 0.765    & 0.580     & 0.791   & 0.630    & 0.635   & 0.574     & 0.930 \\
                                                                           & Single & 0.135 & \textbf{0.928}  & 0.315 & 0.969  & 0.537 & 0.837   & 0.777    & 0.518     & \textbf{0.819}   & 0.557    & 0.405   & 0.473     & \textbf{0.948} \\ \bottomrule
\end{tabular}
}
\end{table*}

\subsection{Affect and Game Context Classification}
For each task, model, and label values for Precision, Recall, and F1 Score are calculated by:
\begin{equation}
    Precision = \frac{TP}{TP+FP}\\
\end{equation}
\begin{equation}
    Recall = \frac{TP}{TP+FN}\\
\end{equation}
\begin{equation}
    F1 = 2 \cdot  \frac{Precision \cdot Recall}{Precision+Recall}
\end{equation}
\looseness-1Where $TP$, `true positive', is the sum of correctly classified data-points for a label, $FP$, `false positive', is the sum of all data points annotated with a different label but classified as this label and $FN$, `false negative', being the sum of all data points with this label that were classified with a different label. These metrics were used because accuracy, e.g. the average correct classifications across a whole data set, is not necessarily the best measure of success when testing on very unbalanced data. In these cases, high accuracy can be achieved by simply always classifying the majority class e.g. for Arousal outputting only Neutral Arousal classifications would yield an accuracy of $0.94$. As such, discussion of the results will focus on the individual class F1 Score values, as it is the harmonic average of Precision and Recall so is representative of both. F1 scores for all models across all classes in both single and joint task learning are shown in Table \ref{table:prf1}. 

\subsection{Joint vs Single Task Learning}
\looseness-1One of the aims of this study is to explore if learning both game context and affect classifications at the same time would improve results. While learning these tasks simultaneously has the drawback that model has fewer variables to dedicate to each task, jointly learning the task may also lead to learning more generalisable, and thus robust, representations. Additionally, joint task learning has the benefit of requiring only a single model to perform recognition across all tasks thus resulting in faster training and inference as only one model needs to be trained/queried.
 
\begin{figure}[t]
\centering
\includegraphics[width=0.5\textwidth]{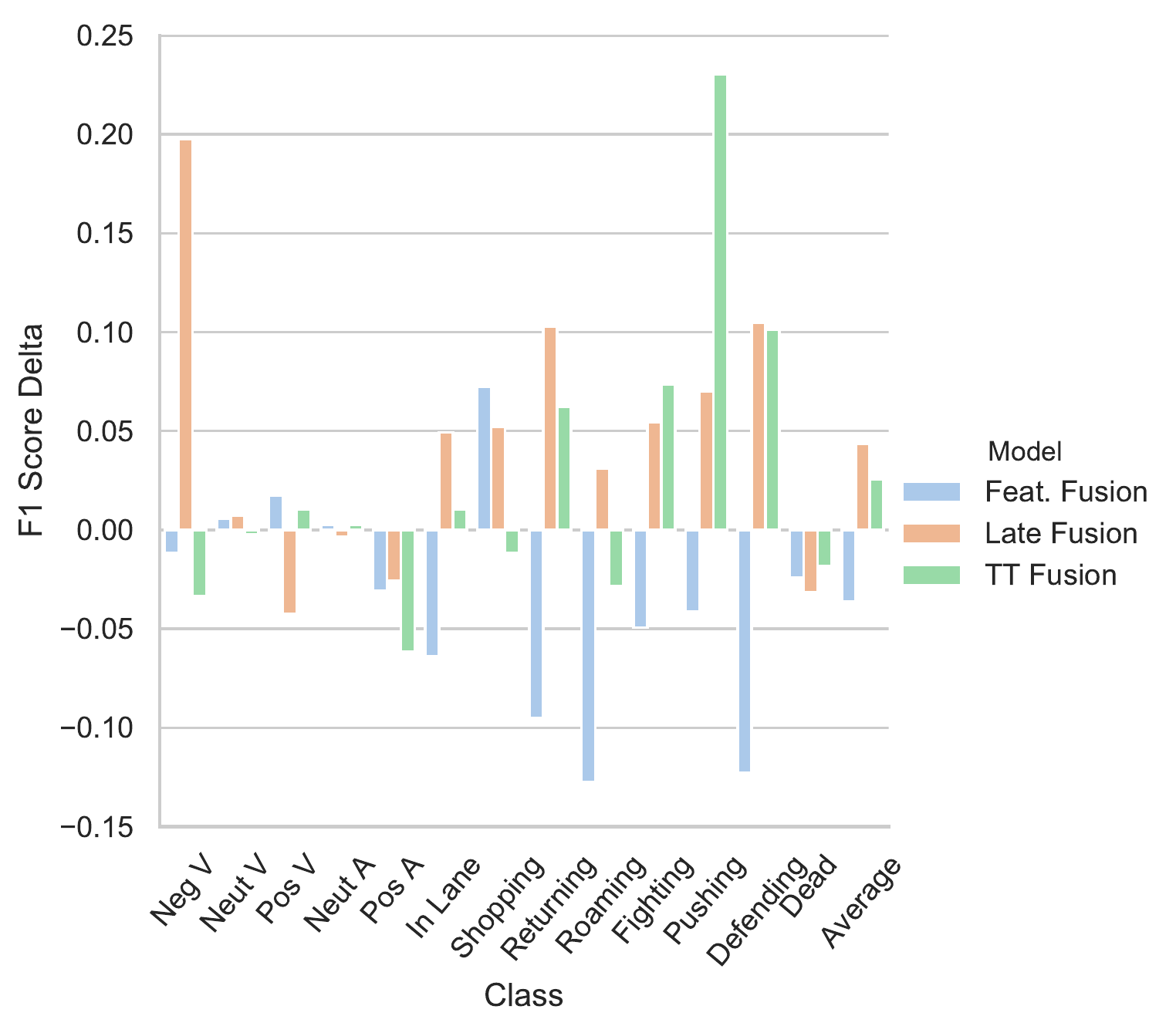}
\caption{Delta change in F1 Score performance. Positive values occur when joint task learning outperforms single task learning. Negative values occur when single task learning outperforms joint task learning.}
\label{fig:deltas}
\end{figure}

\looseness-1Fig. \ref{fig:deltas} shows the delta in performance between single and joint task learning, showing that whilst there is a large degree of between-class variance, the models which perform fusion after the LSTM step see an improvement when learning jointly, whereas the early fusion model performs worse.

\section{Discussion}
\subsection{Affect and Game Context Classification}
\looseness-1Table \ref{table:prf1} shows that affect classification is a much harder task that game context classification, further reinforcing the discussion regarding the difficulty of in-the-wild affect detection. Importantly, while results such as a high of $0.362$ for F1 Score for positive valence and a high of $0.286$ for negative valence may seem poor, they are actually significantly higher than a random baseline, which has expected F1 scores of $0.044$ (negative) and $0.106$ (positive), because of the hugely imbalanced data set. There is clearly still a lot of interesting work to be done to accurately model streamer emotions, however, these results do provide a  baseline. Additionally, Tensor Train models have a large scope for improvement gains via hyper-parameter tuning \cite{Yang2017} so it is possible that future gains are possible using similar architectures.

\looseness-1Regarding game context we see that in general all models perform better when classifying examples of `In Lane', `Shopping', `Roaming' and `Dead', where F1 scores range from $0.667$ (Early Fusion, `Roaming', joint task learning) to $0.948$ (TT Fusion, `Dead', single task learning), compared to other context classes. It is difficult to know exactly why the models perform better on these categories but it is possible confusion arises for classes such as `Pushing', and `Defending' which can be visually similar. Consider the case where there are two players of opposite teams very close to each other and at the edge of a base. Only the positioning of the two players and prior knowledge about which side they are on provide clues as to if the streamer is pushing or defending. Contrast this with the `Dead' class where the screen is mostly greyscale with a fixed message on the screen, resulting in easier classification.

\subsection{Joint vs Single Task Learning}
\looseness-1As Fig. \ref{fig:deltas} shows, for Late Fusion and TT Fusion we see a general if erratic improvement when learning a joint representation with average F1 Score improvements of $0.044$ and $0.026$. However, Early Fusion sees a degradation of F1 Score performance with an average change of $-0.036$. Furthermore by performing a Wilcoxon Signed-Rank Test we see that both the change in Early Fusion and Late Fusion are statistically significant at $P < 0.05$ ($P = 0.047$ and $P = 0.03$ respectively). Perhaps a key take away from these results is that seemingly fusion techniques occurring after temporal modelling see improved results from learning both tasks jointly, but Early Fusion which occurs before the LSTM see a performance reduction. 

\subsection{Determining the `Best' Model}
%It is difficult to ascertain from these results which model performs the best as there is no clear `winner' across all categories. Each model appears to perform better on some tasks and worse on others, and this could very well be an indication that more data is needed. 

\looseness-1It is difficult to ascertain from these results which model performs the best as there is no clear `winner' across all categories. Each model appears to perform better on some tasks and worse on others. Early fusion outperforms TT fusion and late fusion for emotion prediction tasks, although TT fusion appears best at classifying the neutral, majority, classes. This is possibly due to the LSTM layers applied to the fused representation being able to facilitate a temporal, multimodal representation that can account for cross-view cues manifesting at different points in time, e.g. anticipatory co-articulation.  Furthermore, TT fusion slightly outperforms late fusion in emotion recognition due to TT fusion better modelling the interactions between views. In general, using post-LSTM fusion (late, TT-fusion) seems to provide better performance at game context recognition, possibly due to the game footage and game audio being the most important cues for the task, and there appears to be a benefit in modelling their individual dynamics separately.  

The early fusion network is significantly larger in terms of weights than the other two models, in fact, they are roughly only 2/3\textsuperscript{rds} of the size. Additionally, it is the only model that sees performance degradation when joint task learning. Therefore whilst its performance is comparable to other models it does so using approximately 3 million more weights and therefore has a much higher computational cost.

\section{Conclusions and Future Work}
In this work, we pose the problem of modelling streamer affect jointly with game context, within the framework of a deep learning architecture that fuses audio-visual stream data exploiting the power of tensor decompositions such as Tensor Train.  Furthermore, the first annotated data set of emotion and game context for video game livestreams is presented, along with baseline results comparing early and late fusion approaches to the proposed model. Results show that the proposed approach generally outperforms Early Fusion and is comparable to Late Fusion across tasks. Additionally, the difficulty of affect classification in this environment is shown. Clearly, jointly modelling game context and streamer affect in a multimodal setting constitutes a challenging problem of an interdisciplinary nature, with challenges arising in the context of areas such as computer vision, machine learning, and affective computing. Ultimately, this paper acts as a call for action, inviting more researchers to work in this area, as improved results and models are crucial for facilitating audio-visual player experience modelling on livestreams with implications across a range of disciplines.

\bibliographystyle{IEEEtran}
\bibliography{cut_bib}
\end{document}